\documentclass[a4paper, 10pt, conference]{ieeeconf}      
\usepackage{FG2017}
\usepackage{times}
\usepackage{epsfig}
\usepackage{graphicx}
\usepackage{amsmath}
\usepackage{amssymb}
\usepackage{algorithm}
\usepackage{algpseudocode}
\usepackage{booktabs}
\usepackage{multirow}
\usepackage{subfigure}
\usepackage{mathtools}
\usepackage{epstopdf}
\usepackage{boldline}

\newcolumntype{M}[1]{>{\centering\arraybackslash}m{#1}}

\DeclarePairedDelimiter{\norm}{\lVert}{\rVert}

\FGfinalcopy 

\IEEEoverridecommandlockouts                              
\title{\LARGE \bf
A Proximity-Aware Hierarchical Clustering
of Faces }

\author{\parbox{16cm}{\centering
		{\large Wei-An Lin, Jun-Cheng Chen, and Rama Chellappa}\\
		{\large
			University of Maryland, College Park}\\
		{\texttt{walin@terpmail.umd.edu, pullpull@cs.umd.edu, rama@umiacs.umd.edu}}}
}

\begin{document}
\IEEEoverridecommandlockouts\pubid{\makebox[\columnwidth]{978-1-5090-4023-0/17/\$31.00~\copyright{}2017 IEEE \hfill} \hspace{\columnsep}\makebox[\columnwidth]{ }}

\ifFGfinal
\thispagestyle{empty}
\pagestyle{empty}
\else
\author{Anonymous FG 2017 submission\\-- DO NOT DISTRIBUTE --\\}
\pagestyle{plain}
\fi
\maketitle

\begin{abstract}
In this paper, we propose an unsupervised face clustering algorithm called
``Proximity-Aware Hierarchical Clustering'' (PAHC) that exploits the local
structure of deep representations. In the proposed method,
a similarity measure between deep features is computed by evaluating linear SVM margins. SVMs are trained using nearest neighbors of sample data, and thus do not require any
external training data. Clusters are then formed by thresholding the
similarity scores. We evaluate the clustering performance using
three challenging unconstrained face datasets, including Celebrity
in Frontal-Profile (CFP), IARPA JANUS Benchmark A (IJB-A), and JANUS
Challenge Set 3 (JANUS CS3) datasets. Experimental results
demonstrate that the proposed approach can achieve significant
improvements over state-of-the-art methods. Moreover, we also
show that the proposed clustering algorithm can be applied to curate
a set of large-scale and noisy training dataset while maintaining
sufficient amount of images and their variations due to nuisance factors. The face verification
performance on JANUS CS3 improves significantly by finetuning a DCNN
model with the curated MS-Celeb-1M dataset which contains over three
million face images.
\end{abstract}

\section{Introduction} \label{sec:intro}

In this paper, we address the problem of face clustering, especially
for the scenario of grouping a set of face images without knowing
the exact number of clusters. Face clustering algorithms provide
meaningful partitions for given face image sets by combining faces
with similar appearances while separating dissimilar ones. Ideally,
face images in a partition should belong to the same identity, while
images from different partitions should not. Identity-sensitive
face clustering algorithms is an active research in computer vision
and has several applications, including but not limited to
organizing personal pictures, summarizing imagery from social media,
and homeland security camera during
investigation. Clustering is also important when we need large
amount of data to train a deep convolutional neural network (DCNN)
for face verification, classification, or detection tasks. Recently,
Microsoft Research released the MS-Celeb-1M
dataset~\cite{guo_ms_2016}, which contains 1M celebrity names
and over 8 million face images. Due to its diversity, this
very-large dataset has the potential to improve the performance of face
recognition systems. However, since the MS-Celeb-1M dataset has been built from the outputs of search engines, labeling errors could adversely affect the training of deep networks. An effective
approach to tackle this problem is to apply a reliable clustering
algorithm on the MS-Celeb-1M training dataset to harvest sufficient number of face
images that can be used for training a DCNN.

Despite extensive studies on general clustering algorithms over the
past few decades, face image clustering remains a difficult task.
The difficulties are mainly two-fold. Since face images of a person
may have variations in illumination, facial expressions, occlusion,
age, and pose, it is challenging to measure the similarity between two
face images. Another issue is that without knowing the actual number
of clusters, many well-established clustering algorithms, such as $k$-means, may not be effective.

Recent advances on DCNNs have brought about impressive improvements for image
classification and verification tasks~\cite{simonyan_verydeep_2014,
krizhevsky_imagenet_2012}, which can be attributed to its ability to
extract discriminative information from each image and represent it
compactly. Inspired by this progress, we
apply a DCNN to extract deep features from the given face and define a similarity
measure to separate one face from another. Traditional methods define pairwise similarity
based on monotonically decreasing functions of distance, \emph{e.g.} the
Gaussian kernel $\exp (-d(\mathbf{x}_i, \mathbf{x}_j)^2/\sigma^2)$.
Recently, Zhu \emph{et al.}~\cite{Zhu_2011_CVPR} proposed Rank-Order
clustering where pairwise similarity is measured based on the
ranking of shared nearest neighbors. Otto \emph{et
al.}~\cite{Otto_2016_cluster_face} improved the scalability and
accuracy of Rank-Order clustering by considering only the presence
or absence of nearest neighbors. We hypothesize, based on the works by
Zhu and Otto, that neighborhood geometry should be considered to
achieve improved clustering performance. However, Rank-Order
clustering computes similarities based on shared nearest neighbors
in a domain where geometrical information may be lost (\emph{i.e.},
`rank' only contains the ordering information). Our approach
measures the similarity between neighborhoods directly in the feature space: neighborhood geometries are first transferred
to an evaluation hyperplane, pairwise similarity is then defined by
evaluating the points on the hyperplanes.

The rest of the paper is organized as follows. We present our
algorithm in Section~\ref{sec:proposed_approach}. Qualitative and
quantitative evaluations are conducted in Section~\ref{sec:exp}. Finally, conclusions are given in Section~\ref{sec:conclusion}.

\section{Related Work}
\begin{figure*}[!t]
    \centering
    \includegraphics[width=6.5in, keepaspectratio]{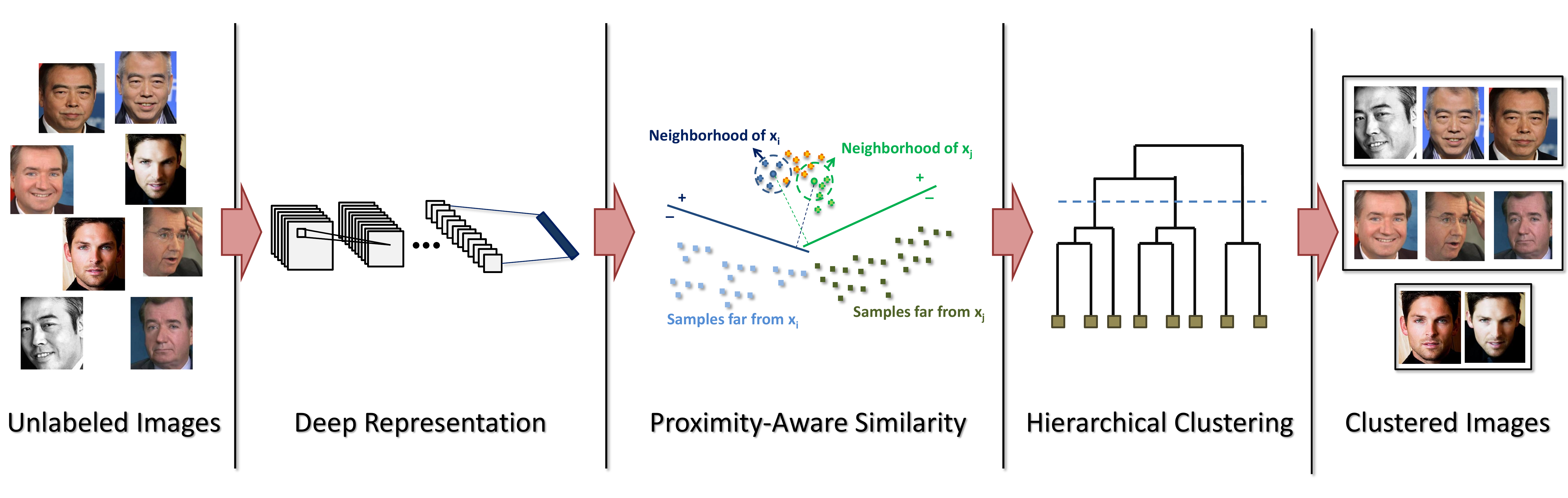}
    \caption{Overall pipeline for the proposed PAHC algorithm. Unlabeled face images are preprocessed and passed through a DCNN to obtain deep features. The Proximity-Aware similarity between each pair of features is then computed. Based on the Proximity-Aware similarity, hierarchical clustering is applied to yield the final image clusters. } \label{fig:pipeline}
\end{figure*}

\textbf{General Clustering Algorithms} Clustering algorithms can be
generally categorized into partitional and agglomerative approaches.
Both approaches build upon a similarity graph $G(V, E)$ defined for the given dataset. The graph can be either fully connected, in
$\epsilon$-neighborhood or in $k$-nearest neighbor. For partitional
approaches, given the number of clusters,
$k$-means~\cite{Macqueen_kmeans} iteratively updates the group centers and
corresponding members until convergence. Spectral
clustering finds the underlying structure based on graph
Laplacian~\cite{laplacian,Ng_spectral,self-tuningspectral}. For
agglomerative
approaches~\cite{Gowda_1978_agglomerative,Kurita_1991_hierarchical},
groups of data points are merged whenever the linkage between them
is above some threshold. Finding the proper similarity measure is one
of the major tasks in designing clustering algorithms. Traditional
approaches use non-increasing functions of pairwise distance as
the similarity measure, \emph{e.g.} $\exp(-d(\mathbf{x}_i, \mathbf{x}_j)^2/\sigma^2)$. Recently, sparse subspace
clustering (SSC)~\cite{Elhamifar_2009_CVPR,Elhamifar_2013} and low-rank subspace clustering (LRSC)~\cite{Vidal_2014,Liu_2011_lowRank}, which
exploit the subspace structures in a dataset, have gained some attention. Both methods assume
data points are self-expressive. By minimizing the reconstruction error
under the sparsity/low-rank criterion, the similarity matrix can be obtained
from the corresponding sparse/low-rank representation. However, SSC
and LRSC are computationally expensive and hard to scale.
In~\cite{Vishal_2015_latentSSC}, dimensionality reduction and
subspace clustering are simultaneously learned to achieve improved
performance and efficiency. Another category, known as
supervised clustering, learns appropriate distance metric from
additional
datasets~\cite{Jordan_2003_learningSpectral,Bar-Hillel_2005_learn_Maha,Finley_2005_SVMClustering,Law_2016_CVPR}.

\textbf{Image Clustering Algorithms} Yang \emph{et
al.}~\cite{Yang_2016_CVPR} proposed learning deep representations
and image clusters jointly in a recurrent framework. Each image is
treated as separate clusters at the beginning, and a deep network is
trained using this initial grouping. Deep representation and cluster
members are then iteratively refined until the number of clusters
reached the predefined value. Zhang \emph{et
al.}~\cite{Zhang_2016_videoClustering} proposed to cluster face
images in videos by alternating between deep representation adaption
and clustering. Temporal and spatial information between and within
video frames is exploited to achieve high purity face image
clusters. Zhu \emph{et al.}~\cite{Zhu_2011_CVPR} measured
pairwise similarity by considering the ranks of shared nearest
neighbors, and transitively merged images into clusters when the
similarity is above some threshold. Otto \emph{et
al.}~\cite{Otto_2016_cluster_face} modified the algorithm by (i) using
deep representations of images (ii) considering only the absence and
presence of the shared nearest neighbors and (iii) transitively merging
only once. Superior clustering results and computational time are
achieved from the modifications. Sankaranarayanan \emph{et
al.}~\cite{Swami_2016_triplet} proposed learning a low-dimensional
discriminative embedding for deep features and applied hierarchical
clustering to realize state-of-the-art precision-recall clustering
performance on the LFW dataset.

Different from these studies, we propose a clustering algorithm that
does not require (i) training a deep network
iteratively~\cite{Yang_2016_CVPR} (ii) partial identity
information~\cite{Zhang_2016_videoClustering} and (iii) additional
training data~\cite{Swami_2016_triplet}. Our approach focuses on
exploiting the neighborhood structure between samples and implicitly performing
domain adaptation to achieve improved clustering performance.

\section{Proposed Approach} \label{sec:proposed_approach}
In this section, we introduce our clustering algorithm, illustrated in Fig.~\ref{fig:pipeline}. The face images first
pass through a pre-trained face DCNN model to extract the deep features.
Then, we compute the Proximity-Aware similarity scores using linear
SVMs trained with corresponding neighborhoods of the samples.
Finally, the agglomerative hierarchical clustering method is applied on the
similarity scores to determine the cluster labels to each sample.
The details of each components are described in the following
subsections.

\subsection{Notation}
We denote the set of face images as $I = \{I_1, \ldots, I_{n_s}\}$.
Our goal is to assign labels $L = \{l_1, \ldots, l_{n_s}\}$ for each
image to indicate the cluster it belongs to. The images are first
passed through a pre-trained DCNN model to extract the deep features,
which are then normalized to unit length. Specifically, let
$f_{\theta}: \mathcal{I} \rightarrow \mathcal{X}$ be the DCNN
network parameterized by $\theta$, and $g: \mathcal{X} \rightarrow
\mathcal{X}$ be the normalization. The corresponding deep
representations for the face images are given by $X = g \circ
f_{\theta}(I) = \{\mathbf{x}_1, \ldots, \mathbf{x}_{n_s}\}$. For each representation
$\mathbf{x}_i$, we define $\mathcal{N}_K(\mathbf{x}_i)$ as the set of $K$-nearest
neighbors of $\mathbf{x}_i$, including $\mathbf{x}_i$ itself.

\subsection{Agglomerative Hierarchical Clustering}
Agglomerative hierarchical
clustering~\cite{Gowda_1978_agglomerative,Kurita_1991_hierarchical}
initializes all samples as separate clusters. Based on the pairwise distance matrix $\mathbf{D}$ measured from the features, clusters are iteratively merged whenever the cluster-to-cluster distance is
below some threshold $\eta$. The hierarchical clustering
algorithm, denoted as $\mathrm{Hierarchical}(\mathbf{D}, \eta)$, generates the cluster assignments $L$ for all the faces in $I$. In our work, we use average
linkage as a measure of cluster-to-cluster distance.

Recent advances in DCNN have yielded great improvements for face verification task, which uses cosine distance as the similarity measure to decide whether two faces belong to the same subject. Given two features $\mathbf{x}_i, \mathbf{x}_j \in \mathcal{X}$ on the unit hypersphere $\{\mathbf{x} : \norm{\mathbf{x}} = 1 \}$, the similarity measure between them is computed by
\begin{equation}
s(\mathbf{x}_i, \mathbf{x}_j) = \mathbf{x}_i^T\mathbf{x}_j. \label{eq:sim-original}
\end{equation}
The pairwise distance matrix $\mathbf{D}$ in this case is simply
\begin{equation}
[\mathbf{D}]_{i,j} = 1 - s(\mathbf{x}_i, \mathbf{x}_j). \label{eq:d_matrix}
\end{equation}
Since DCNNs trained on large datasets extract discriminative features for images, $\mathrm{Hierarchical}(\mathbf{D}, \eta)$, where $\mathbf{D}$ is from~\eqref{eq:d_matrix}, can perform well on datasets that have similar distribution as the training dataset. However, the difference in distribution encountered in many real-world applications degrades the performance significantly. Inspired by previous works~\cite{Zhu_2011_CVPR,Otto_2016_cluster_face}, we aim at measuring similarity based on the neighborhood structure.

\subsection{Proximity-Aware Similarity}

To have a formulation that is able to take neighborhoods
$\mathcal{N}_K(\mathbf{x}_i)$, $\mathcal{N}_K(\mathbf{x}_j)$ into account when
measuring the similarity between $\mathbf{x}_i$ and $\mathbf{x}_j$, we rewrite the inner
product as
\begin{equation}
s(\mathbf{x}_i, \mathbf{x}_j) = \dfrac{\mathbf{x}_i^T\mathbf{x}_j + \mathbf{x}_j^T\mathbf{x}_i}{2}. \label{eq:sim-proposed}
\end{equation}
In~\eqref{eq:sim-proposed}, the similarity between $\mathbf{x}_i$ and $\mathbf{x}_j$ is evaluated by averaging two asymmetric measures: How similar is $\mathbf{x}_j$ from the view of $\mathbf{x}_i$ and how similar is $\mathbf{x}_i$ from the view of $\mathbf{x}_j$. Specifically, $\mathbf{x}_i^T\mathbf{x}_j$ can be interpreted as evaluating $\mathbf{x}_j$ on hyperplane $H_i = \{ \mathbf{x}: \mathbf{x}_i^T \mathbf{x} = 0\}$ and $\mathbf{x}_j^T\mathbf{x}_i$ can be interpreted as evaluating $\mathbf{x}_i$ on hyperplane $H_j = \{ \mathbf{x}: \mathbf{x}_j^T \mathbf{x} = 0\}$. This observation allows us to generalize the asymmetric measure as follows.

\begin{figure}[!t]
    \centering
    \includegraphics[width=3.0in]{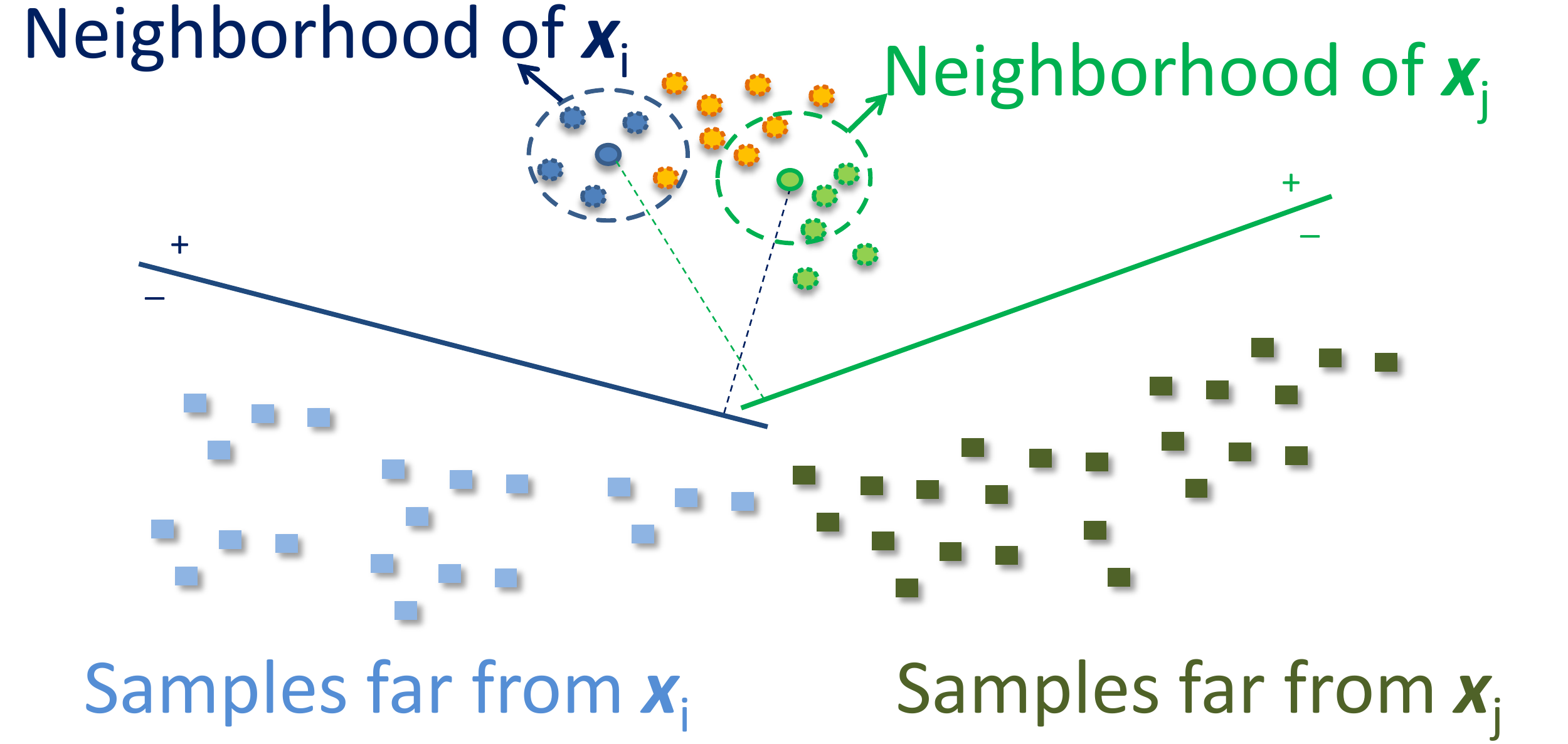}
    \caption{Proximity-Aware similarity. Circles in blue, yellow, green represent samples with different identities. Blue dashed circles delineate the neighborhood of $\mathbf{x}_i$ (or $\mathcal{N}_6(\mathbf{x}_i)$) while green dashed circles delineate the neighborhood of $\mathbf{x}_j$ (or $\mathcal{N}_6(\mathbf{x}_j)$). The blue hyperplane is obtained by solving~\eqref{eq:svm}, treating $\mathcal{N}_K(\mathbf{x}_i)$ as positive samples, and a subset of $X \backslash \mathcal{N}_K(\mathbf{x}_i)$, which are blue squares in this case, as negative samples. The green hyperplane is obtained in the same way. The Proximity-Aware similarity between $\mathbf{x}_i$ and $\mathbf{x}_j$ is evaluated using~\eqref{eq:sim-neighbor}. The length of the blue dashed line and the green dashed line reflects how similar are the two neighborhoods.} \label{fig:large_margin}
\end{figure}

Given a hyperplane $H_{\mathbf{w}_i, b_i} = \{\mathbf{x}: \mathbf{w}_i^T\mathbf{x} + b_i = 0\}$ which contains information about $\mathcal{N}_K(\mathbf{x}_i)$, the asymmetric similarity from $H_{\mathbf{w}_i, b_i}$ to some set $S$ is defined as
\begin{equation}
H_{\mathbf{w}_i, b_i}(S) = \dfrac{1}{|S|}\sum_{\mathbf{x} \in S} [\mathbf{w}_i^T\mathbf{x} + b_i ].
\end{equation}
Following~\eqref{eq:sim-proposed}, the generalized similarity measure, which we call ``Proximity-Aware similarity'', is the average of two asymmetric measures from $H_{\mathbf{w}_i, b_i}$ to $\mathcal{N}_K(\mathbf{x}_j)$ and from $H_{\mathbf{w}_j, b_j}$ to $\mathcal{N}_K(\mathbf{x}_i)$:
\begin{equation}
s_{PA}(\mathbf{x}_i, \mathbf{x}_j) = \dfrac{H_{\mathbf{w}_i, b_i}(\mathcal{N}_K(\mathbf{x}_j)) + H_{\mathbf{w}_j, b_j}(\mathcal{N}_K(\mathbf{x}_i))}{2}. \label{eq:sim-neighbor}
\end{equation}
Unlike cosine similarity, $s_{PA}$ is not bounded. We introduce a nonlinear transformation to define the Proximity-Aware pairwise distance
\begin{equation}
[\mathbf{D}_{PA}]_{ij} = 1 - \dfrac{2}{\pi}\arctan\left[ s_{PA}(\mathbf{x}_i, \mathbf{x}_j) \right].
\end{equation}
This choice of nonlinearity is for experimental simplicity. One can also consider $[\mathbf{D}_{PA}]_{i,j} = \exp(-s_{PA}(\mathbf{x}_i, \mathbf{x}_j))$. The Proximity-Aware Hierarchical clustering is then characterized by the following algorithm:
\begin{equation}
L_{PA} \leftarrow \mathrm{Hierarchical}(\mathbf{D}_{PA}, \eta).
\end{equation}
The above construction helps us to cast the problem of defining the similarity function between neighborhoods into finding hyperplanes $H_{\mathbf{w}_i, b_i}$. Our ultimate goal is to find a similarity measure for each pair of feature vectors that reflects whether they belong to the same class. We conjecture that $H_{\mathbf{w}_i, b_i}$ and $H_{\mathbf{w}_j, b_j}$ should have the following property:
\\ \\
\emph{$H_{\mathbf{w}_i, b_i}(\cdot)$ has a large value when evaluating on sets that are near $\mathcal{N}_K(\mathbf{x}_i)$, and has a small value otherwise.}
\\ \\
\noindent The constraint not only forces the similarity measure to be locally geometry-sensitive (proximity-aware) but also adaptive to the data domain. This motivates the use of linear classifiers to separate positive samples $\mathcal{N}_K(\mathbf{x}_i)$ from their corresponding negative samples. Fig.~\ref{fig:large_margin} shows a demonstrative example. This approach is analogous to the one-shot similarity technique~\cite{Lior_2009_ICCV}. In this work, we use the linear SVM as our candidate algorithm for finding hyperplanes. Specifically, we solve

\begin{align}
\min_\mathbf{u} \dfrac{1}{2} \mathbf{u}^T \mathbf{u} + C_p \sum_{k = 1}^{N_p} \max[0, 1-y_k\mathbf{u}^T \mathbf{z}_k]^2 \nonumber \\
+ C_n \sum_{k=1}^{N_n} \max[0, 1-y_k\mathbf{u}^T \mathbf{z}_k]^2, \label{eq:svm}
\end{align}
where $\mathbf{u} = [\mathbf{w}^T \quad b]^T$ and $\mathbf{z}_k = [\mathbf{x}_k^T \quad 1]^T$. We treat $\mathcal{N}_K(\mathbf{x}_i)$ as positive samples with cardinality $N_p$, and a subset of $X \backslash N_K(\mathbf{x}_i)$ as negative samples with cardinality $N_n$. $y_k = +1$ for positive samples and $y_k = -1$ for negative samples. The regularization constants $C_p$ and $C_n$ are given by $C_p = C \frac{N_p + N_n}{N_p}$ and $C_n = C \frac{N_p + N_n}{N_n}$.

In~\cite{Lior_2009_ICCV}, Linear Discriminant Analysis (LDA) is used as the
classifier to evaluate one-shot similarity score. However, we do not consider LDA as our candidate because
the bimodal Gaussian prior assumption is not always satisfied for
the positive and negative samples drawn from real-world datasets. In
the proposed method, negative samples often consist of features from different identities with variations from nuisance factors, which do not obey a single Gaussian distribution.  

\subsection{Choice of Positive and Negative Sets} \label{sec:choice}

Since the hyperplane is chosen based on large-margin classification
between positive samples $\mathcal{N}_K(\mathbf{x}_i)$ and negative
samples, the choice of them would be crucial. In this paper, we
first construct the nearest neighbor list $\mathrm{NNList}_{\mathbf{x}_i}$ for
each data sample $\mathbf{x}_i$, where $\mathrm{NNList}_{\mathbf{x}_i}[1] = \mathbf{x}_i$.
$\mathcal{N}_K(\mathbf{x}_i)$ corresponds to
$\mathrm{NNList}_{\mathbf{x}_i}[1:K]$, and we choose
$\mathrm{NNList}_{\mathbf{x}_i}[N_1:N_2]$ as the negative samples. In
Section~\ref{sec:exp}, we show how parameters $(K, N_1, N_2)$ affect
the clustering performance in detail.

\section{Experimental Results} \label{sec:exp}
In this section, we evaluate our clustering algorithm qualitatively
on the recently released MS-Celeb-1M~\cite{guo_ms_2016} dataset
and quantitatively on the IARPA Janus Benchmark-A (IJB-A), JANUS
Challenge Set 3 (CS3), and Celebrities in Frontal-Profile (CFP)
datasets. To compute Proximity-Aware similarity,
we use the LIBLINEAR library~\cite{lib_linear} with L2-regularized L2-loss primal SVM. The parameter $C$ is set at $10$ throughout this section.\\
\\
\textbf{MS-Celeb-1M~\cite{guo_ms_2016}:}

Microsoft Research recently released this very large face image
dataset, consisting of 1M identities. The training dataset of
MS-Celeb-1M is prepared by selecting top 99,892 identities from the
1M celebrity list. There are 8,456,240 images in total, roughly 85
images per identity. This dataset is designed by leveraging a
knowledge base called ``freebase''. Since face images are created using a search engine, labeling noise may be
a problem when this dataset is used in supervised learning tasks. We demonstrate the effectiveness of the proposed clustering algorithm in curating large-scale noisy dataset in Section~\ref{sec:exp_quality} and Section~\ref{sec:finetune}. In this paper, we directly use the aligned images provided along with the dataset. \\
\\
\textbf{Celebrities in Frontal-Profile (CFP)~\cite{sengupta_frontal_2016}:}

\begin{figure}[!h]
    \centering
    \includegraphics[width=2.3in, keepaspectratio]{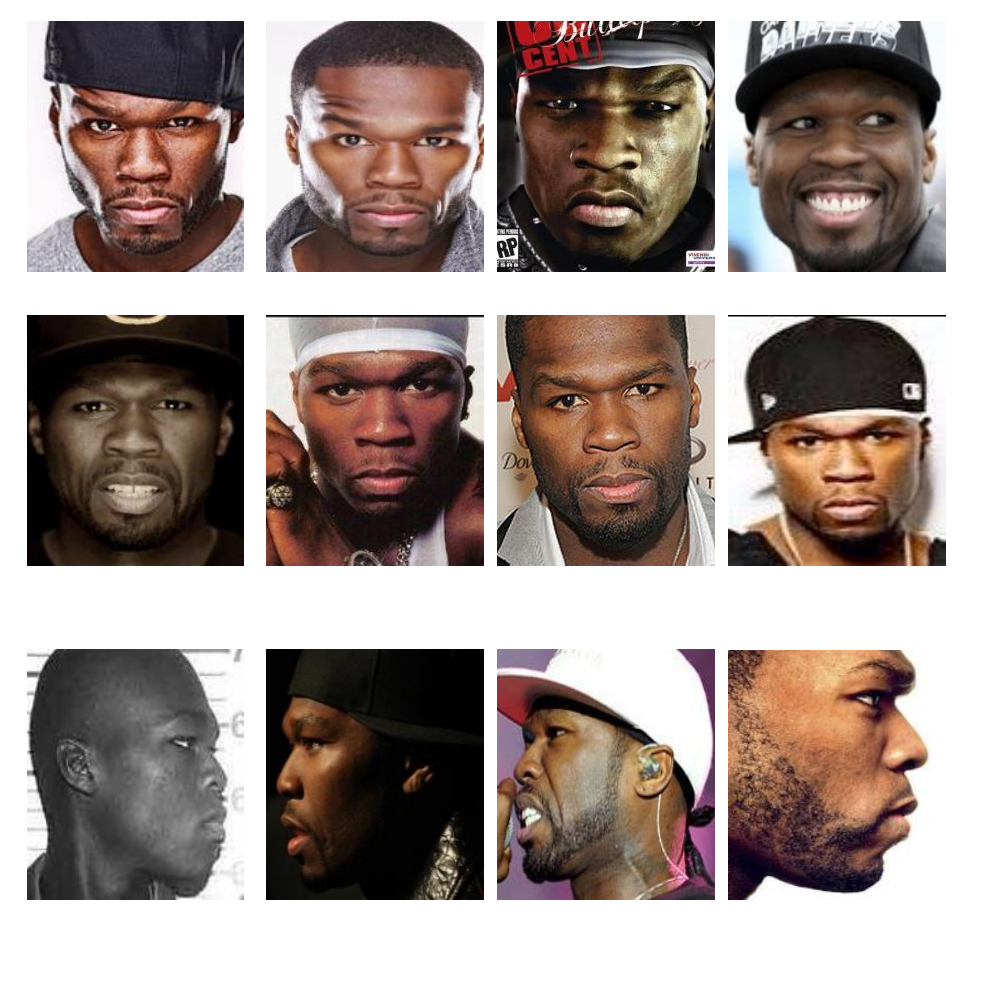}
    \caption{Sample images in CFP dataset. The first two rows are frontal face images and the last row consists of profile face images.} \label{fig:POSE_sample}
\end{figure}

This dataset contains 500 subjects and 7,000 face images. Of the 7,000 faces, 5,000 are in frontal view, and the remaining 2,000 are in profile views where each subject contains 10 frontal and 4 profile
images. Unlike the IJB-A dataset, the CFP dataset aims at isolating the
factor of pose variation in order to facilitate research in
frontal-profile face verification. Extreme variations in poses can be
seen in Fig.~\ref{fig:POSE_sample}. In this work, we apply our
clustering algorithm on all 7,000 face images. \\
\\
\textbf{IARPA Janus Benchmark A (IJB-A)~\cite{klare_janus_2015} and JANUS Challenge Set 3 (CS3):}

\begin{figure}[!h]
    \centering
    \includegraphics[width=2.3in, keepaspectratio]{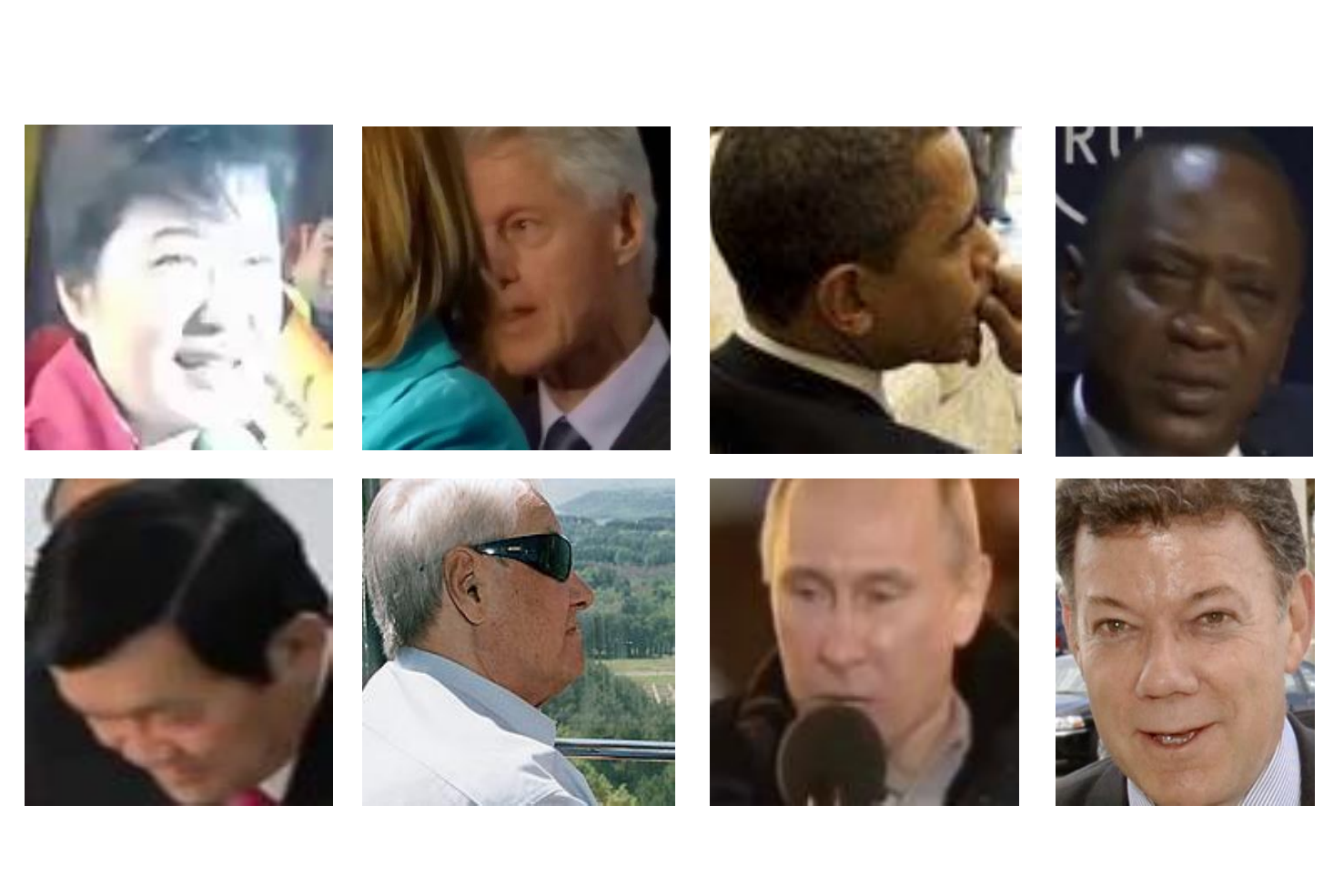}
    \caption{Sample images in CS3 dataset. The faces contain extreme illumination, viewpoint, pose, and occlusion changes.} \label{fig:CS3_sample}
\end{figure}

The IJB-A dataset contains 500 subjects with a total of 25,813 images
taken from photos and video frames (5,399 still images and 20,414
video frames). Extreme variations in illumination, resolution,
viewpoint, pose and occlusion make it a very challenging dataset.
In this work, we cluster the templates corresponding to the query set for each split in IJB-A 1:1 verification protocol
where a template is composed of a combination of still images and video frames. The CS3 dataset is a superset of IJB-A dataset which
contains 11,876 still images and 55,372 video frames sampled from 7,094 videos. In this paper, we use images and video frames provided in CS3 1:1 verification protocol. There are totally 1,871 subjects and 12,590 templates. Fig.~\ref{fig:CS3_sample} shows sample images from different templates.\\

\subsection{Preprocessing} \label{sec:preprocess}
All the face bounding boxes and fiducial points
for the IJB-A, and JANUS CS3 datasests are extracted using
the multi-task face and fiducial detector of Hyperface
\cite{ranjan_hyperface_2016}. For the MS-Celeb-1M and CFP datasets, we use the fiducial points or aligned images provided with the
dataset. Each face is aligned into the canonical coordinate with the
similarity transform using seven landmark points: two
left eye corners, two right eye corners, nose tip, and two mouth
corners. For the CFP dataset, since fiducial points are only
available for half face, we use two eye corners of one of eyes, nose
tip, and one of mouth corners. In training phase, data augmentation is performed by randomly cropping and horizontally flipping face images.

\subsection{Deep Network and Image Representation} \label{sec:deep}

We implement the network architecture presented in
\cite{chen_unconstrained_2016} and train it using the CASIA-WebFace
dataset~\cite{yi_learning_2014}. We preprocess this dataset using the steps presented in Section~\ref{sec:preprocess}. We denote this pretrained network
as `DCNN$_{face}$ (CASIA)'. This network is further finetuned with
curated MS-Celeb-1M dataset \cite{guo_ms_2016}, which we denote as `DCNN$_{face}$
(CASIA+MSCeleb)'. The process of removing
mislabeled images in MS-Celeb-1M will be introduced in the next
section. DCNN$_{face}$ (CASIA) is trained using SGD for 780K iterations with a standard batch size 128 and momentum 0.9. The learning rate is set to 1e-2 initially and is halved every 100K iterations. The weight decay rates of all the
convolutional layers are set to 0, and the weight decay of the final
fully connected layer is set to 5e-4. Pretraining the DCNN network
with the CASIA dataset not only provides good initialization for
the model parameters but also greatly reduces the training time on
the curated MS-Celeb-1M dataset. Then, we finetune the pretrained network to obtain DCNN$_{face}$
(CASIA+MSCeleb) for improved face representation. We use the learning rate 1e-4 for all the convolutional layers, and 1e-2 for the fully connected layers. The network is trained for 240K iterations. In the training phase of DCNN$_{face}$ (CASIA) and DCNN$_{face}$
(CASIA+MSCeleb), the dropout ratio is set as 0.4 to
regularize fc6 due to the large number of parameters (\emph{i.e.}
320 $\times$ 10503 for the CASIA dataset and 320 $\times$ 58207 for the curated MS-Celeb-1M dataset.). Note that we manually remove the overlapping subjects with the IJB-A and
JANUS CS3 datasets from the CASIA-WebFace and the MS-Celeb-1M datasets. 

The inputs to the networks are $100 \times 100 \times 3$ RGB images. Given a face image, deep representation is extracted from the pool5 layer with dimension 320. In the case of IJB-A and CS3 datasets, if there are multiple images and frames in one template, we perform media average pooling to produce the final representation.

\subsection{Qualitative Study on MS-Celeb-1M} \label{sec:exp_quality}

\begin{figure}[!t]
    \centering
    \subfigure{\includegraphics[width=3.5in, keepaspectratio]{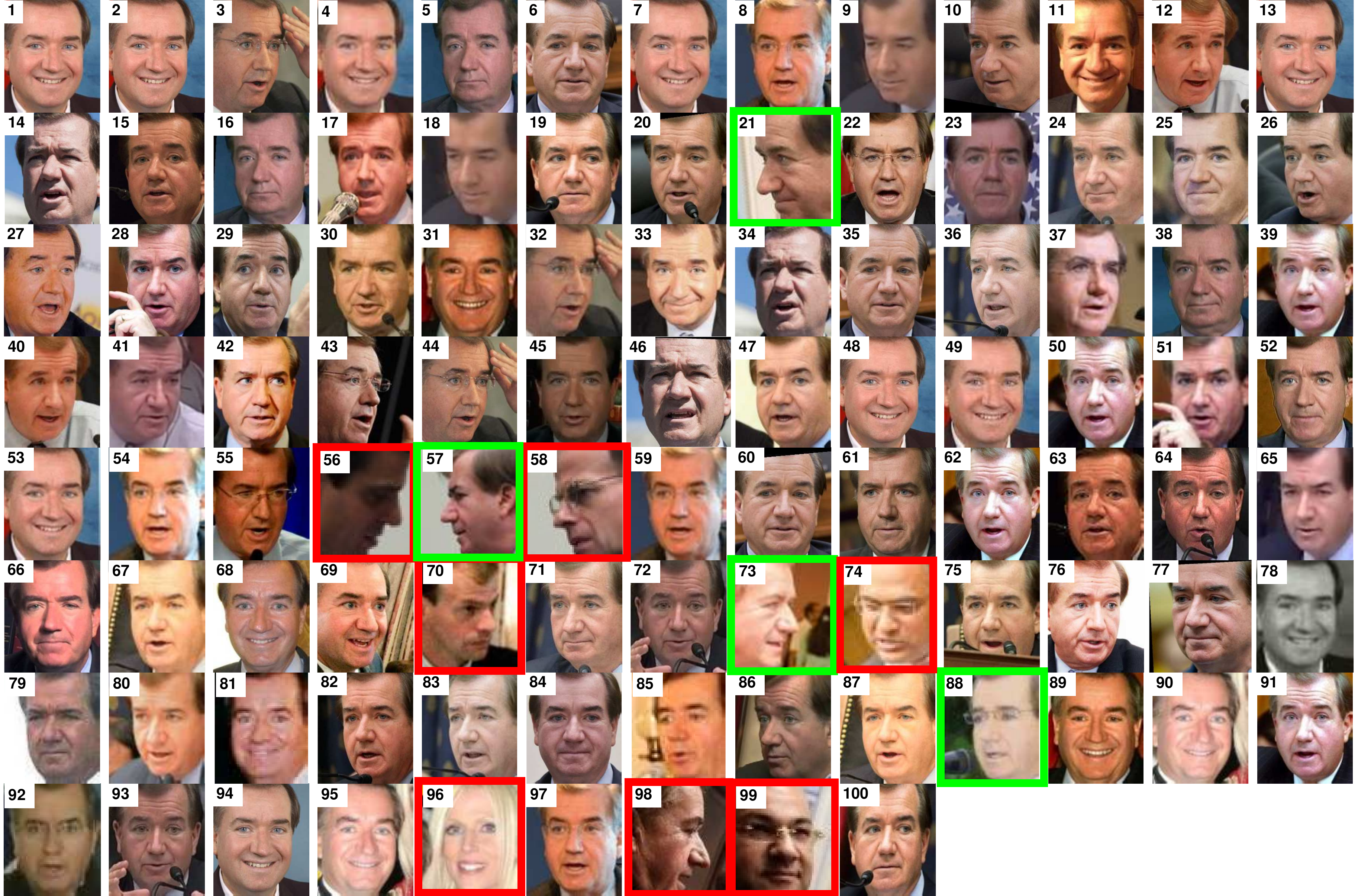}}
    \subfigure{\includegraphics[width=3.5in, keepaspectratio]{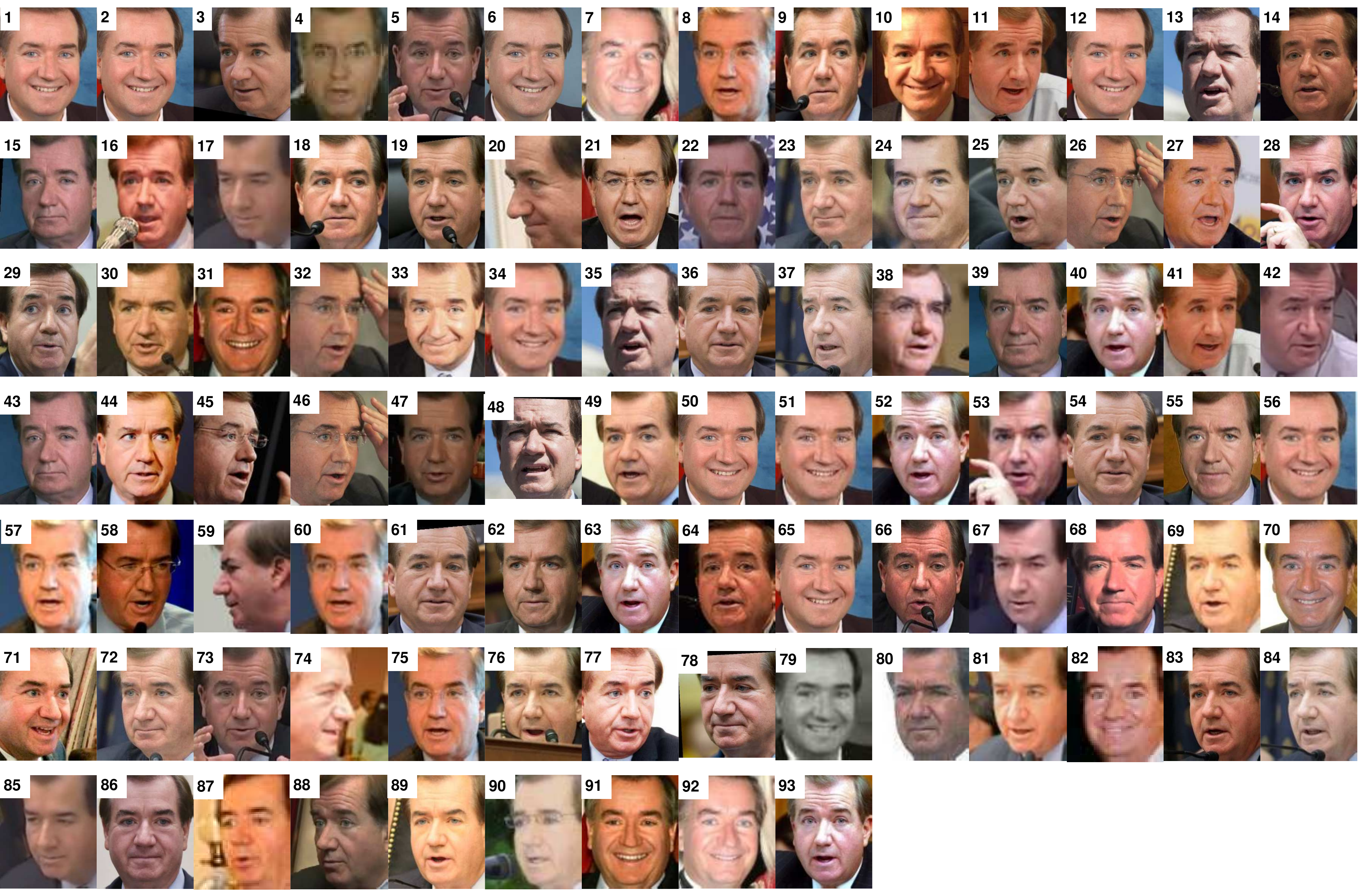}}
    \caption{Sample face images in the MS-Celeb-1M dataset with improved purity after applying the PAHC. Upper-half of the figure shows original face images having machine identifier m.024xcy in MS-Celeb-1M dataset. The lower half of the figure is obtained following the process described in Fig.~\ref{sec:exp_quality}. The red boxes are the face images removed by our algorithm. The green boxes are face images that are retained by our algorithm. The variation in extreme pose (\emph{e.g.} 21, 57, 73) and resolution (\emph{e.g.} 88) will assist the DCNN to learn improved representation.} \label{fig:ms_cleaned}
\end{figure}

As a qualitative study, we apply the clustering algorithm to remove
face images with noisy labels in MS-Celeb-1M training dataset.
Feature representation is first obtained by passing the whole
dataset through DCNN$_{face}$(CASIA) described in
Section~\ref{sec:deep}. We divide the total 99,892 identities into
batches with size 50. For each batch, we apply
$\mathrm{Hierarchical}(\mathbf{D}_{PA}, 2.3)$, with
$[\mathbf{D}_{PA}]_{i,j} = \exp(-s_{PA}(\mathbf{x}_i, \mathbf{x}_j))$. Clusters whose
majority identity have less than 30 images are discarded. The number of the curated dataset is about 3.5
millions face images of 58,207 subjects.
Fig.~\ref{fig:ms_cleaned} shows one example of the clustering results.
Since the PAHC exploits local
property, face images with extreme pose are not discarded. In
addition, our approach does not require external training dataset.

\subsection{Quantitative Study on the CFP, IJB-A, and CS3 datasets}

Images in CFP, IJB-A, and CS3 datasets are first processed as described
in Section~\ref{sec:preprocess}. In this section, we aim to compare our clustering
algorithm with traditional hierarchical clustering, $k$-means, and
Approximate Rank-Order clustering~\cite{Otto_2016_cluster_face}, on the three datasets described earlier. Throughout our experiments, `Approximate Rank-Order clustering' refers to our implementation of the algorithm proposed in~\cite{Otto_2016_cluster_face}. We use the precision-recall curve
defined in~\cite{Otto_2016_cluster_face} as the performance metric
to compare different algorithms at all operation points.
\emph{Pairwise precision} is the fraction of the number of pairs
within the same cluster which are of the same class, over the total
number of same-cluster pairs. \emph{Pairwise recall} is the fraction
of the number of pairs within a class which are placed in the same
cluster, over the total number of same-class pairs.

\begin{figure}[!t]
    \centering
    \includegraphics[width=3.2in, keepaspectratio]{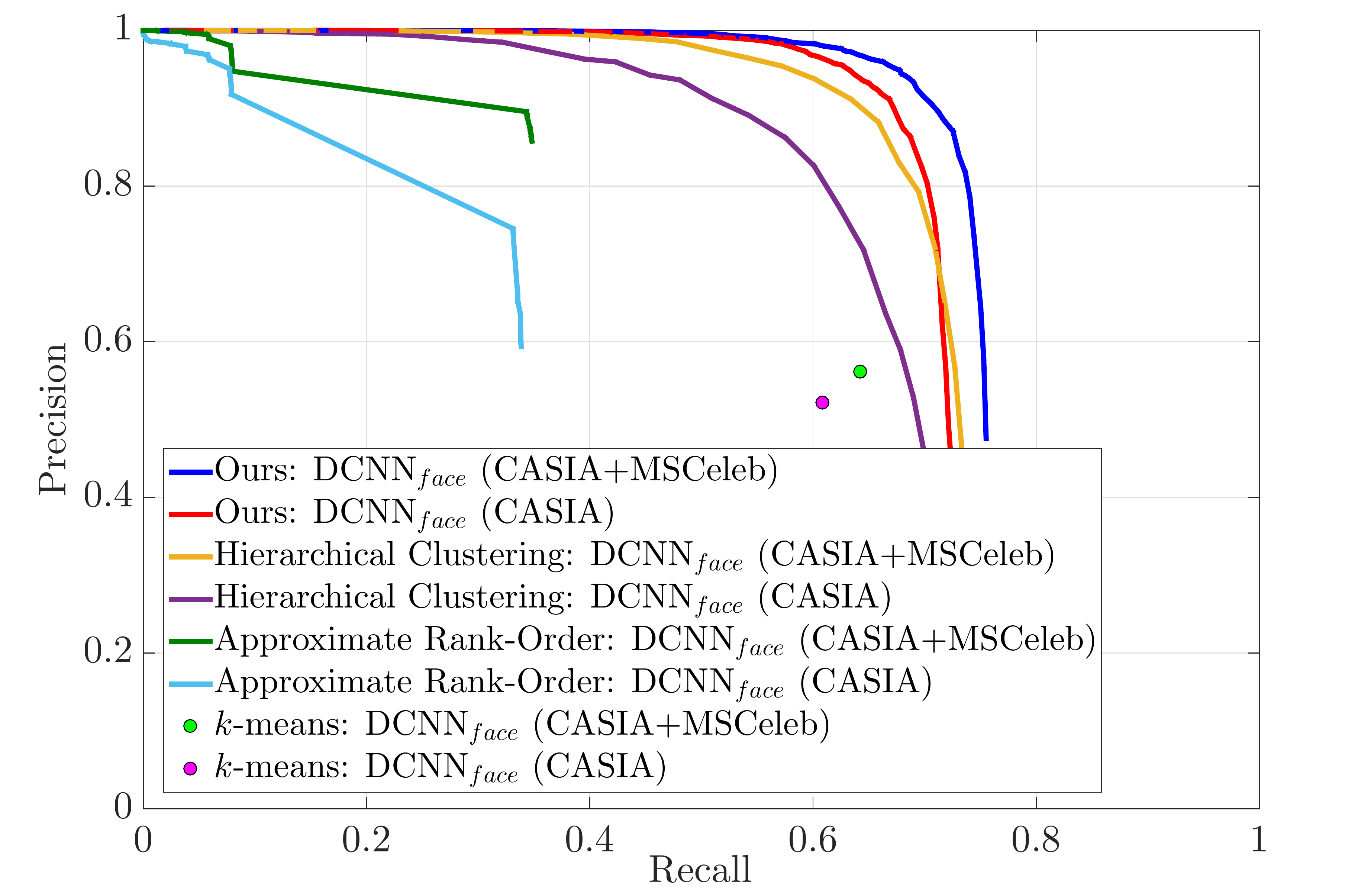}
    \caption{Precision-Recall curve evaluated on the CFP dataset. $(K, N_1, N_2) = (5, 50, 100)$ for the PAHC algorithm, where the parameters are as defined in Section~\ref{sec:choice}} \label{fig:POSE}
\end{figure}

\begin{figure}[!t]
    \centering
    \includegraphics[width=3.2in, keepaspectratio]{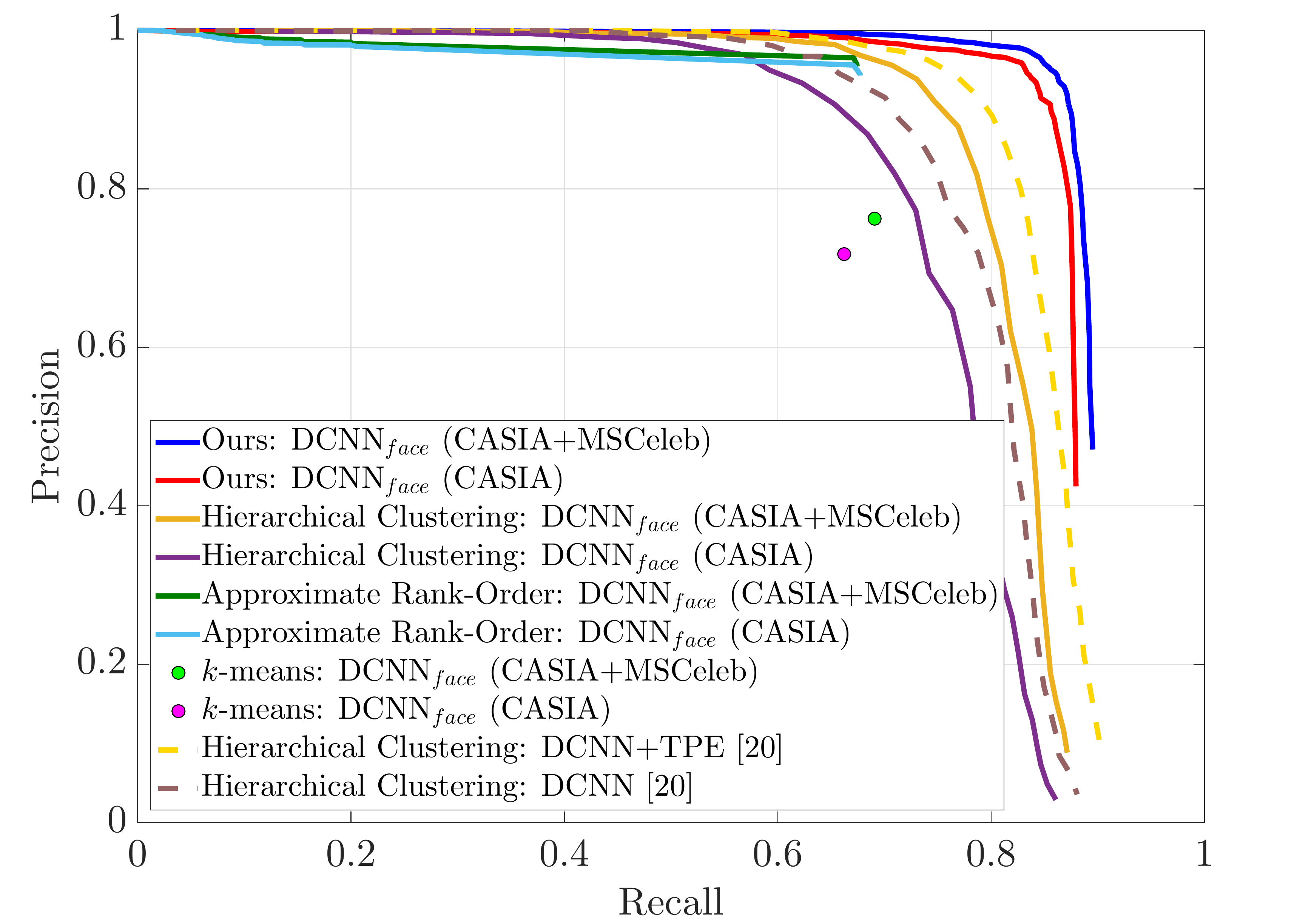}
    \caption{Precision-Recall curve evaluated on the IJB-A dataset. $(K, N_1, N_2) = (5, 50, 100)$ for the PAHC algorithm, where the parameters are as defined in Section~\ref{sec:choice}. It can be observed that the PAHC algorithm outperforms TPE in~\cite{Swami_2016_triplet} by a large margin without the need of external training dataset.} \label{fig:IJBA}
\end{figure}

\begin{figure}[!t]
    \centering
    \includegraphics[width=3.2in, keepaspectratio]{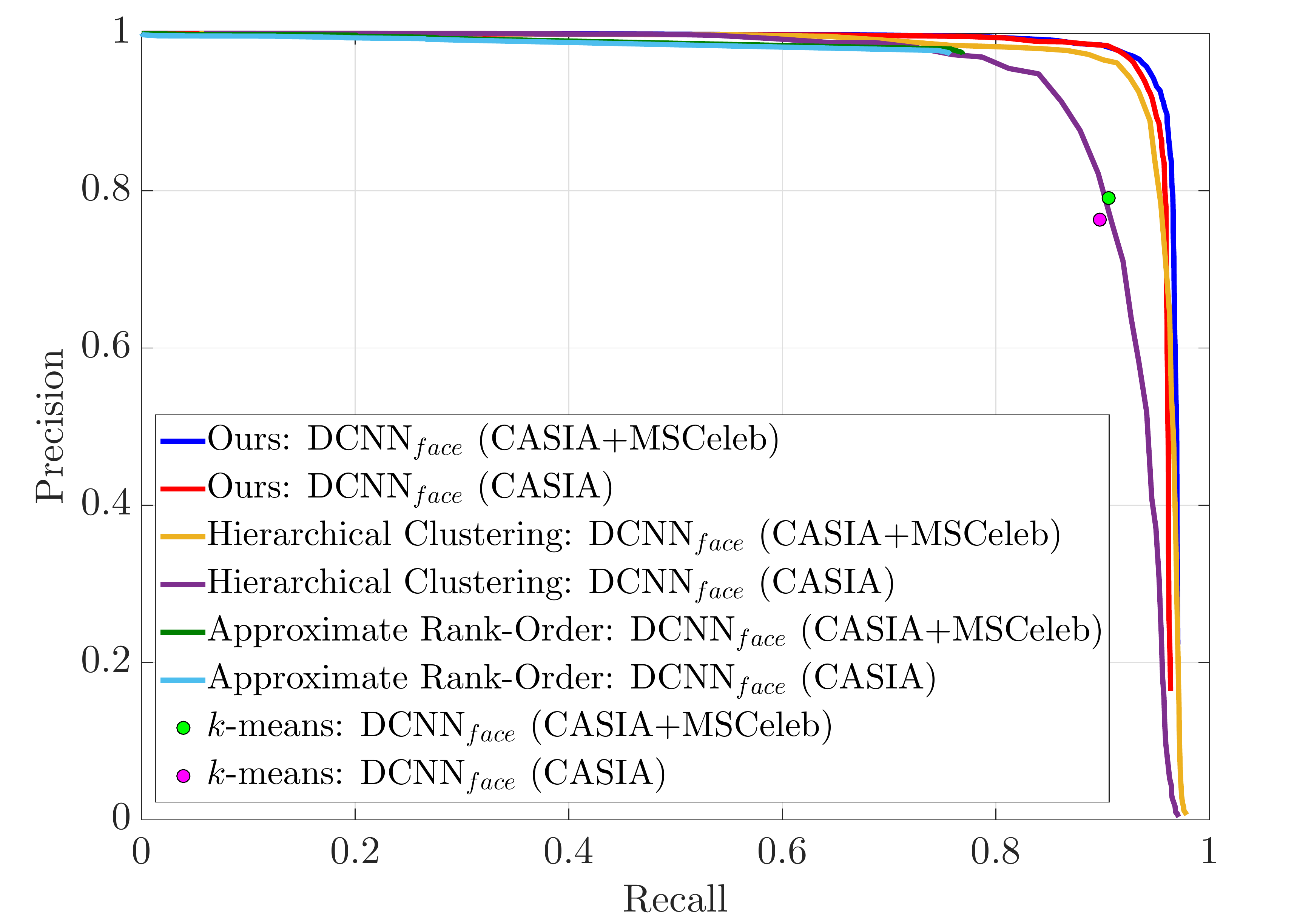}
    \caption{Precision-Recall curve evaluated on the CS3 dataset. $(K, N_1, N_2) = (5, 50, 100)$ for the PAHC algorithm, where the parameters are as defined in Section~\ref{sec:choice}.} \label{fig:CS3}
\end{figure}

\begin{figure}[!t]
    \centering
    \includegraphics[width=2.3in, keepaspectratio]{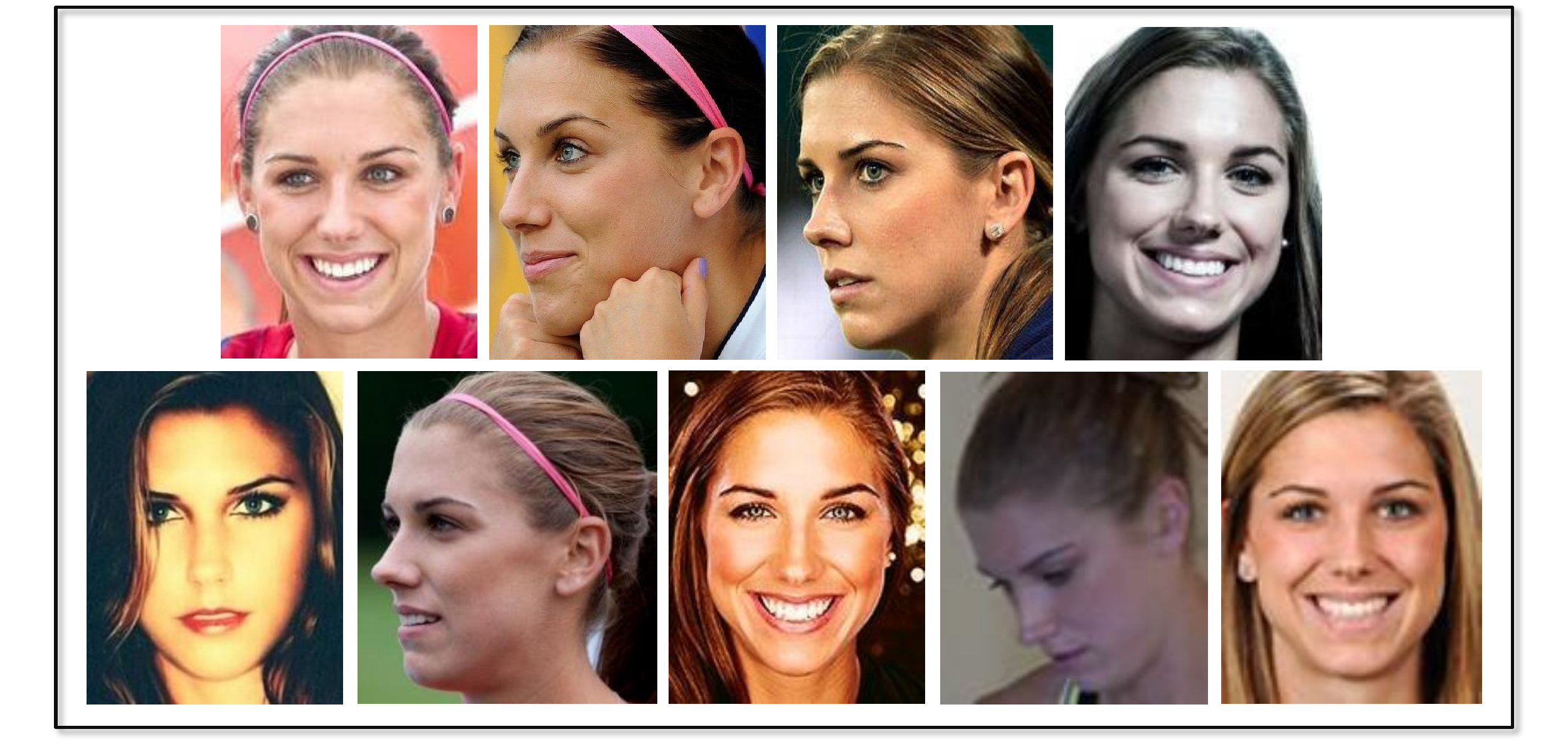}
    \caption{One sample cluster for the CFP dataset after applying the PAHC algorithm.} \label{fig:POSE_cluster}
\end{figure}

\begin{figure}[!t]
    \centering
    \includegraphics[width=2.3in, keepaspectratio]{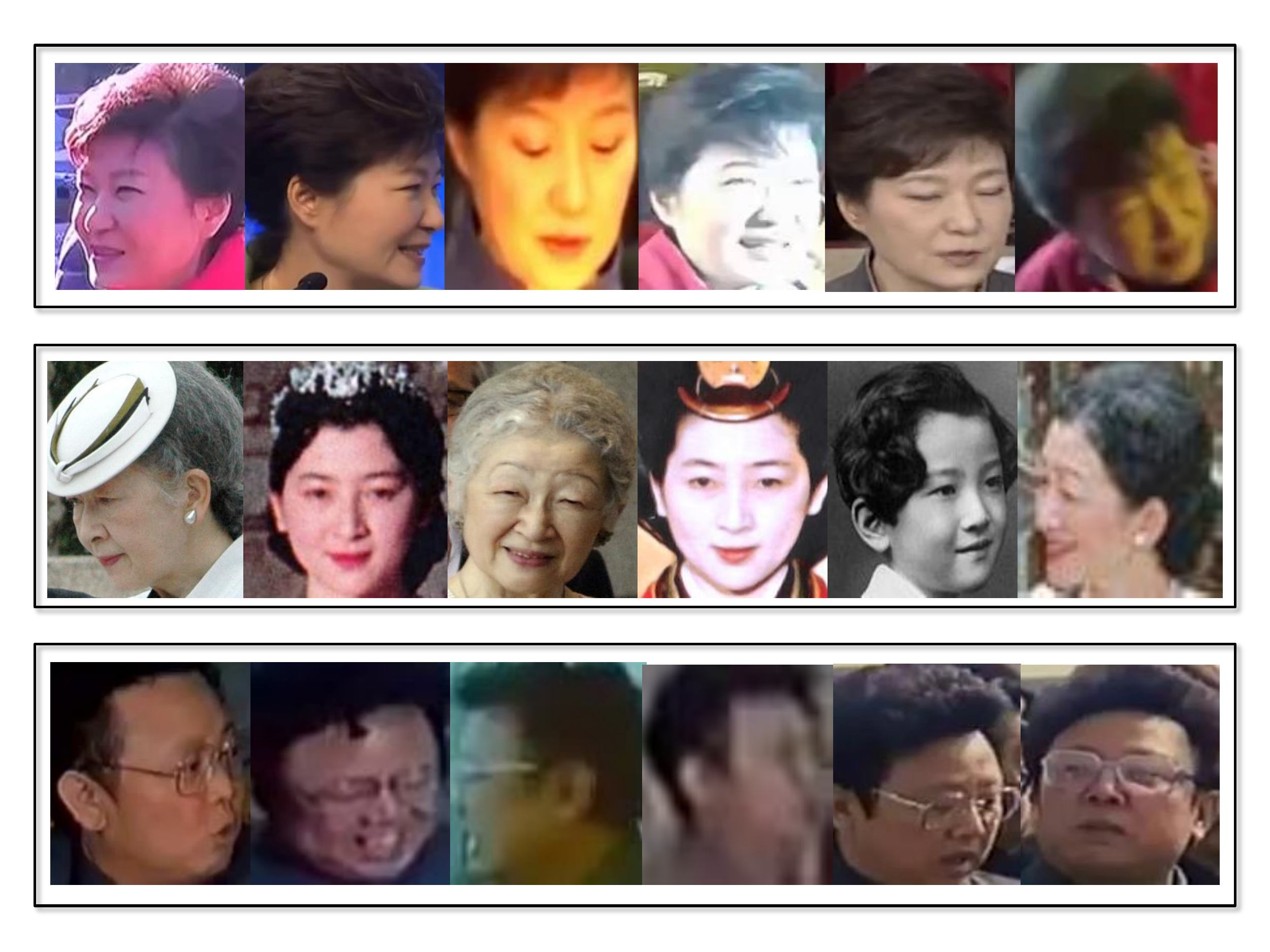}
    \caption{Sample clusters for the CS3 dataset after applying the PAHC algorithm. Robustness to pose variation can be seen throughout the images. Top row shows robustness to illumination changes. Middle row shows robustness to age and makeup. Bottom row shows robustness to blur and viewpoint changes.} \label{fig:CS3_cluster}
\end{figure}

For the CFP dataset, we cluster 7,000 images. For the IJB-A dataset, we cluster the query set provided in the 1:1 verification
protocol for each split, and compute the average performance over 10
splits. For the CS3 dataset, we apply the clustering algorithms on 10,718 probe templates. We use the standard
MATLAB implementation for hierarchical clustering and $k$-means, where
we choose $k$ as the true number of identities. Fig.~\ref{fig:POSE}, Fig.~\ref{fig:IJBA}, and Fig.~\ref{fig:CS3} show the precision-recall performance comparisons, and Fig.~\ref{fig:POSE_cluster} and Fig,~\ref{fig:CS3_cluster} show some sample clustering results.

Compared to hierarchical clustering based on cosine distance, the PAHC attains significant gains by
exploiting the neighborhood structure for each sample. The Approximate
Rank-Order clustering cannot reach recall greater than 0.7 because
it \emph{only computes distances between samples which share a
nearest neighbor}, which means there are some samples which will not be
merged for any choice of thresholding.

\subsection{Parameter and Negative Set Study}

\begin{figure}[!t]
    \centering
    \includegraphics[width=3.2in, keepaspectratio]{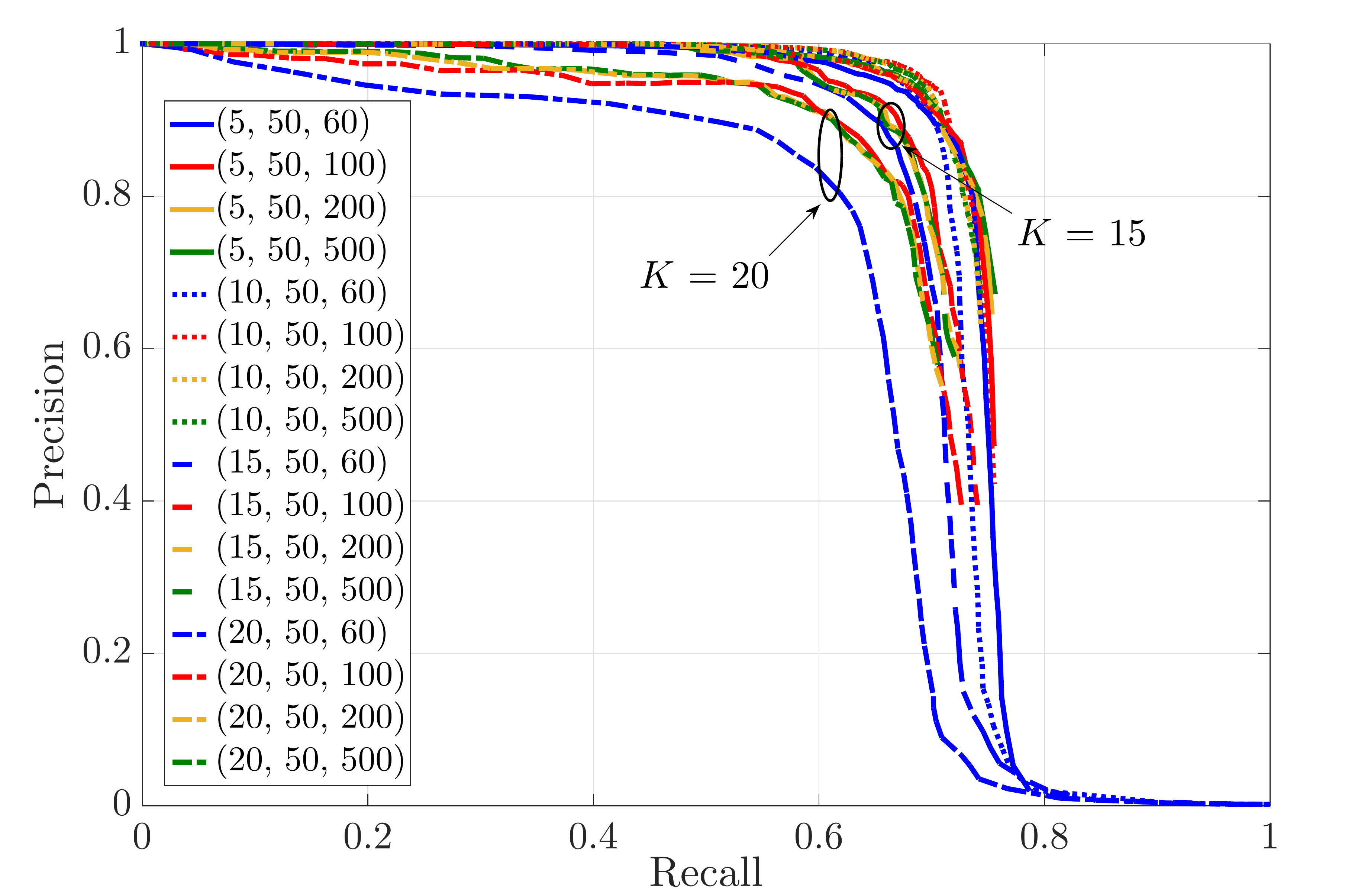}
    \caption{Precision-Recall curves evaluated on the CFP dataset using DCNN$_{face}$ (CASIA+MSCeleb) as the image feature extractor. Results are reported by varying $(K, N_1, N_2)$, where the parameters are as defined in Section~\ref{sec:choice}.} \label{fig:POSE_para_ms}
\end{figure}

\begin{figure}[!t]
    \centering
    \includegraphics[width=3.2in, keepaspectratio]{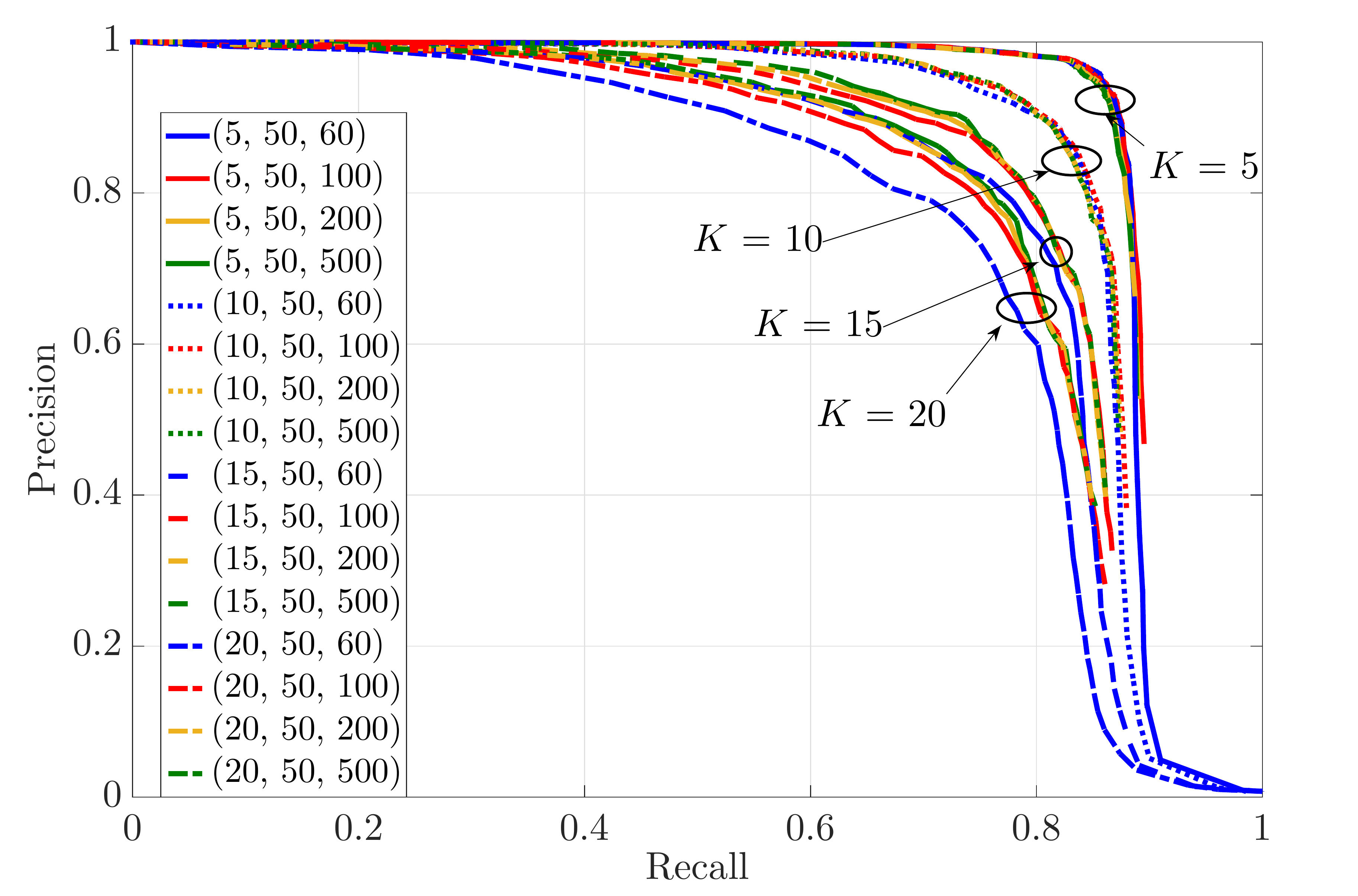}
    \caption{Precision-Recall curves evaluated on the IJB-A dataset using DCNN$_{face}$ (CASIA+MSCeleb) as the image feature extractor. Results are reported by varying $(K, N_1, N_2)$, where the parameters are as defined in Section~\ref{sec:choice}.} \label{fig:IJBA_para_ms}
\end{figure}

\begin{figure}[!t]
    \centering
    \includegraphics[width=3.2in, keepaspectratio]{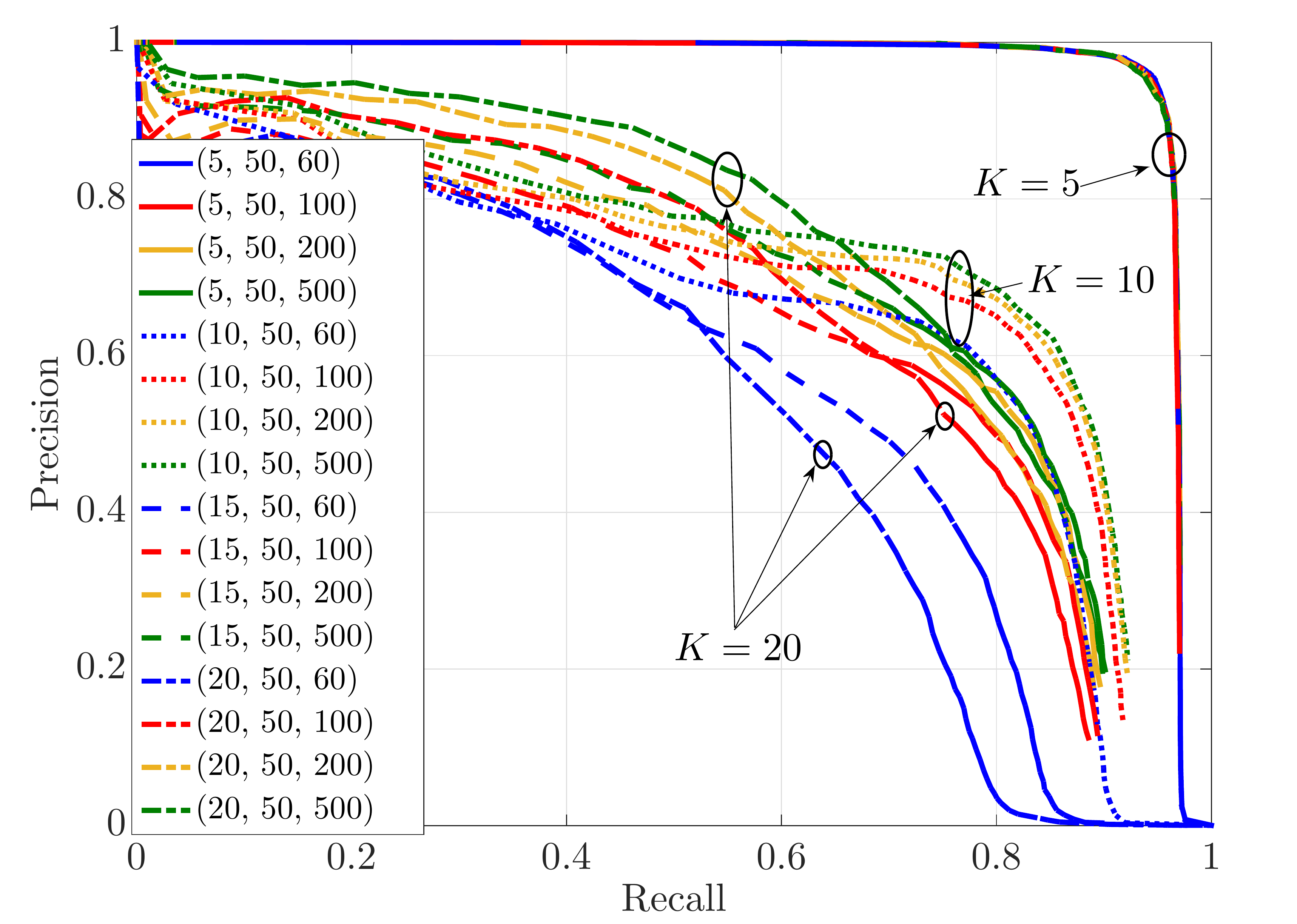}
    \caption{Precision-Recall curves evaluated on the CS3 dataset using DCNN$_{face}$ (CASIA+MSCeleb) as the image feature extractor. Results are reported by varying $(K, N_1, N_2)$, where the parameters are as defined in Section~\ref{sec:choice}.} \label{fig:CS3_para_ms}
\end{figure}

Fig.~\ref{fig:POSE_para_ms},~\ref{fig:IJBA_para_ms}, and~\ref{fig:CS3_para_ms} show results for different parameters settings $(K, N_1, N_2)$ of the proposed algorithm on the CFP, IJB-A, and CS3 datasets. For smaller values of neighborhood size $K$, \emph{e.g.} $K=5$, the choice of negative sets have little effect on the performance. This is because $\mathcal{N}_K(\mathbf{x}_i)$ may not be able to represent the `local' structure for large $K$. In this case, the similarity between $\mathcal{N}_K(\mathbf{x}_i)$ and $\mathcal{N}_K(\mathbf{x}_j)$ would deviate significantly from the similarity between $\mathbf{x}_i$ and $\mathbf{x}_j$. According to our experiments, setting $K = 5$ gives the best performance. 

\begin{figure}[!t]
    \centering
    \includegraphics[width=3.2in, keepaspectratio]{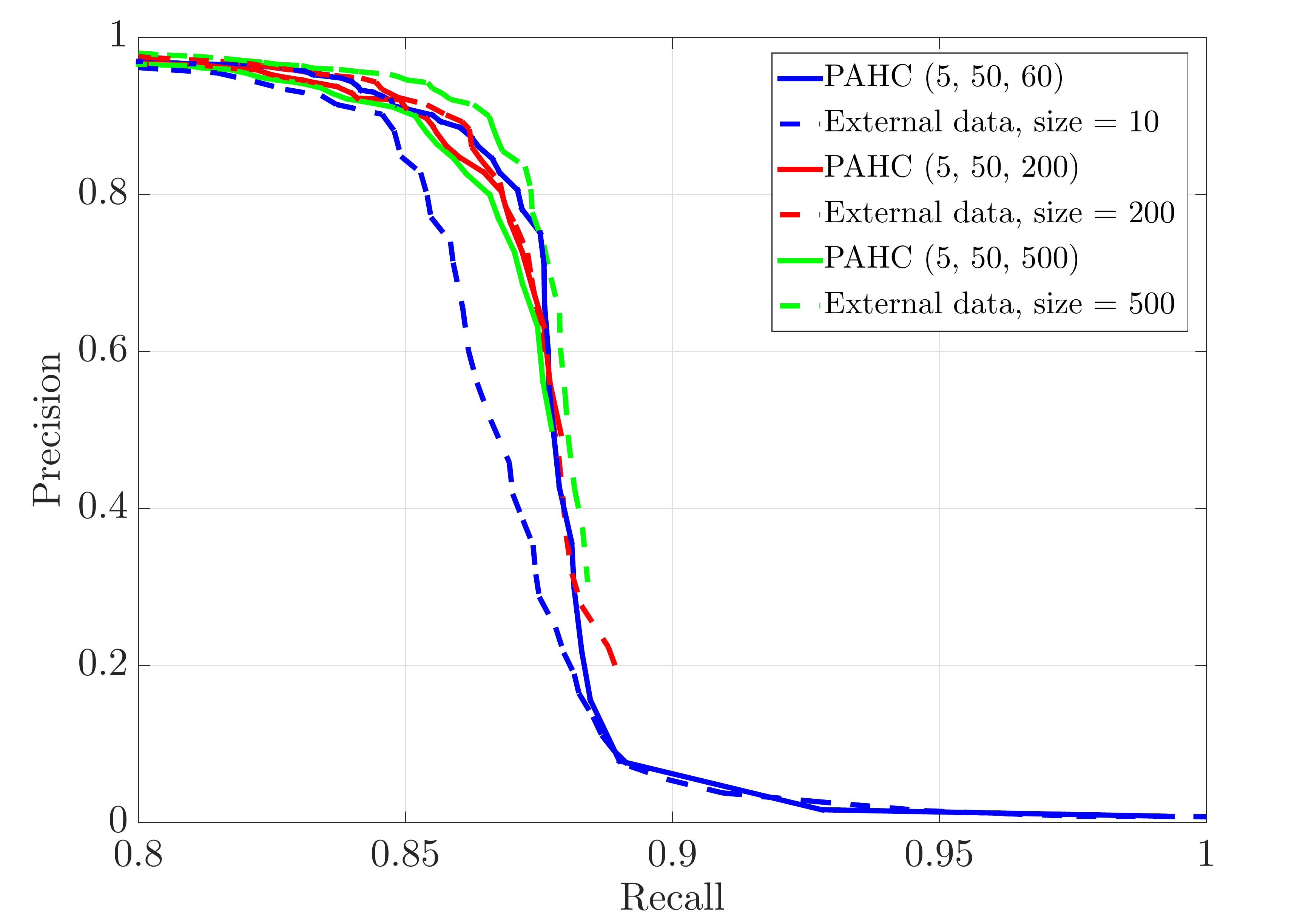}
    \caption{Precision-Recall curves evaluated on the IJB-A dataset. Solid lines indicate the performance of the PAHC for different parameters $(K, N_1, N_2)$. Dashed lines indicate the performance when using a subset of training data as negative samples in a linear SVM.} \label{fig:IJBA_baseline}
\end{figure}

Since we claim that by choosing $\mathrm{NNList}_{\mathbf{x}_i}[N_1:N_2]$ as negative samples, the deep representation can be adapted to the target domain, in the following, we experiment with the IJB-A dataset by sampling templates from the training data in each split and use them as negative samples when training the linear SVM. It can be observed from Fig.~\ref{fig:IJBA_baseline} that when properly labeled templates, which contain nonoverlapping identities with the verification dataset, are used, improved performance is achieved. However, it is not an effective approach to real-world problems since preparing data with nonoverlapping identity with the unseen dataset $I$ is difficult.

\subsection{Finetuning DCNN using Curated MS-Celeb-1M dataset}
\label{sec:finetune}

\begin{figure}[!t]
    \centering
    \includegraphics[width=3.2in, keepaspectratio]{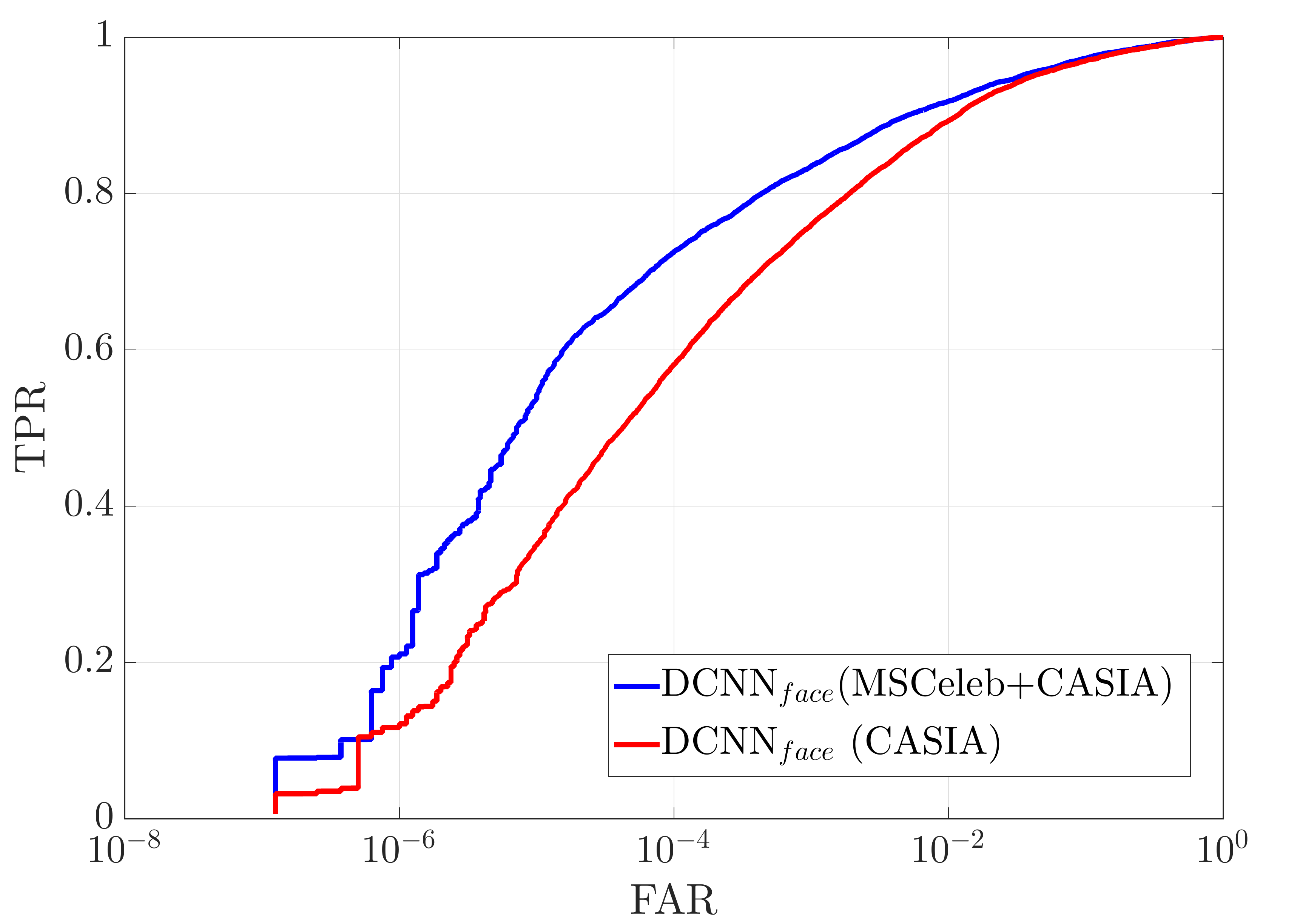}
    \caption{CS3 1:1 verification performance using DCNN$_{face}$ trained on CASIA and CASIA+MSCeleb.} \label{fig:CS3_1:1}
\end{figure}

As described in Section~\ref{sec:deep}, we finetune the pre-trained
DCNN$_{face}$(CASIA) model using the curated subset of MS-Celeb-1M
attained by our clustering algorithm, DCNN$_{face}$(CASIA+MSCeleb).
In contrast, if we do not perform clustering and finetune
DCNN$_{face}$(CASIA) using all the images of MS-Celeb-1M, the model
does not converge. Then, we compare the results of JANUS CS3 1:1
verification task for the two networks: DCNN$_{face}$ (CASIA) and
DCNN$_{face}$ (CASIA+MSCeleb). From Fig.~\ref{fig:CS3_1:1} and Fig.~\ref{table:verification},
DCNN$_{face}$ (CASIA+MSCeleb) outperforms DCNN$_{face}$ (CASIA), and
it demonstrates that the proposed clustering algorithm improves the quality of training data used for the DCNN.

\begin{table}[H]
    \caption{CS3 1:1 verification performance} \label{table:verification}
    \begin{center}
      \begin{tabular}{ c  M{2cm}  M{2.5cm} }
        \hlineB{2}
        FAR & DCNN$_{face}$ \newline (CASIA) & DCNN$_{face}$ \newline (CASIA+MSCeleb) \\  \hline
        \\ [-1em]
        1e-1 & 0.9703 & 0.9731 \\
        \\ [-1em]
        1e-2 & 0.8934 & 0.9184 \\
        \\ [-1em]
        1e-3 & 0.7599 & 0.8355 \\
        \\ [-1em]
        1e-4 & 0.5813 & 0.7252 \\
        \hlineB{2}
      \end{tabular}
    \end{center}
\end{table}

\section{Conclusion} \label{sec:conclusion}

We proposed an unsupervised algorithm, namely, the PAHC algorithm, to measure the pairwise
similarity between samples by exploiting their neighborhood structures
along with domain adaptation. From extensive experiments, we show
that our clustering algorithm achieves high precision-recall performance at all operation points
when the neighborhood is properly chosen. Following this, the PAHC is applied to curate the MS-Celeb-1M training dataset. Our algorithm retains faces with variations in pose, illumination and resolution, while separating images with different identities. We further finetuned the DCNN network with the curated dataset. Significant improvement on CS3 1:1
verification task demonstrates the effectiveness of our algorithm.

\section{Acknowledgments}
This research is based upon work supported by the Office of the
Director of National Intelligence (ODNI), Intelligence Advanced
Research Projects Activity (IARPA), via IARPA R\&D Contract No.
2014-14071600012. The views and conclusions contained herein are
those of the authors and should not be interpreted as necessarily
representing the official policies or endorsements, either expressed
or implied, of the ODNI, IARPA, or the U.S. Government. The U.S.
Government is authorized to reproduce and distribute reprints for
Governmental purposes notwithstanding any copyright annotation
thereon.

{\small
    \bibliographystyle{ieee}
    \bibliography{refs_pullpull,refs_andy}
}

\end{document}